\documentclass[10pt,twocolumn,letterpaper]{article}

\usepackage{cvpr}              
\usepackage[T1]{fontenc}
\usepackage[T1]{fontenc}

\definecolor{cvprblue}{rgb}{0.21,0.49,0.74}
\usepackage[pagebackref,breaklinks,colorlinks,allcolors=cvprblue]{hyperref}
\usepackage{graphicx}
\usepackage{cuted}
\usepackage{capt-of}
\usepackage{blindtext}
\usepackage{graphicx}
\usepackage{caption}
\usepackage{overpic}
\usepackage{appendix}

\usepackage{xcolor}

\usepackage{tikz}
\definecolor{customRed}{RGB}{60,62,245}
\definecolor{customGreen}{RGB}{170,255,172}
\definecolor{customBlue}{RGB}{218,205,93}
\definecolor{customYellow}{RGB}{124,194,255}
\definecolor{customMagenta}{RGB}{255,168,254}
\definecolor{customCyan}{RGB}{210,218,221}

\definecolor{planecolor}{RGB}{31,119,180}
\definecolor{cylindercolor}{RGB}{255,127,14}
\definecolor{conecolor}{RGB}{44,160,44}
\definecolor{spherecolor}{RGB}{214,39,40}
\definecolor{toruscolor}{RGB}{148,103,189}
\definecolor{featurecolor}{RGB}{0,255,0}

\usepackage{times}

\newcommand{\tikzcircle}[2][red,fill=red]{\tikz[baseline=-0.5ex]\draw[#1,radius=#2] (0,0) circle ;}

\def\paperID{9657}
\def\confName{CVPR}
\def\confYear{2025}

\title{CADDreamer: CAD Object Generation from Single-view Images}
\author{%
\large Yuan Li\textsuperscript{1} \quad 
Cheng Lin\textsuperscript{2\dag} \quad 
Yuan Liu\textsuperscript{3} \quad 
Xiaoxiao Long\textsuperscript{4} \quad 
Chenxu Zhang\textsuperscript{5} \quad \\
Ningna Wang\textsuperscript{1} \quad
Xin Li\textsuperscript{6} \quad 
Wenping Wang\textsuperscript{6} \quad 
Xiaohu Guo\textsuperscript{1\dag} \\
\small\textsuperscript{1}University of Texas at Dallas \quad 
\textsuperscript{2}The University of Hong Kong \quad 
\textsuperscript{3}Hong Kong University of Science and Technology \\ 
\small\textsuperscript{4}Nanjing University \quad 
\textsuperscript{5}ByteDance \quad 
\textsuperscript{6}Texas A\&M University \\
\small\dag Corresponding Authors%
}

\begin{document}

\maketitle

\begin{strip}
   \begin{minipage}{\textwidth}\centering
   \vspace{-10pt}
   \captionsetup{type=figure}
   \includegraphics[width=\linewidth,trim=0.4cm 0.4cm 0.4cm 0.4cm, clip]{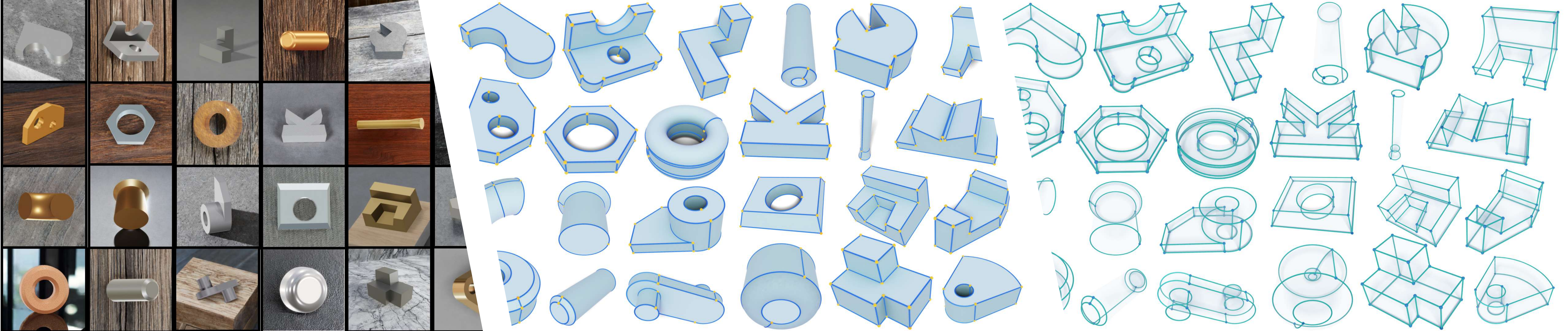}
   \captionof{figure}{A gallery of our reconstructed CAD models (middle and right) from single-view RGB images (left).
    The reconstructed shapes are shown in light blue (middle), while their topological representations of B-rep vertices and edges are shown in green (right).}
   \label{figurelabel}
   \end{minipage}
\end{strip}

\begin{abstract}
    Diffusion-based 3D generation has made remarkable progress in recent years. However, existing 3D generative models often produce overly dense and unstructured meshes, which stand in stark contrast to the compact, structured, and sharply-edged Computer-Aided Design (CAD) models crafted by human designers. To address this gap, we introduce CADDreamer, a novel approach for generating boundary representations (B-rep) of CAD objects from a single image. 
    CADDreamer employs a primitive-aware multi-view diffusion model that captures both local geometric details and high-level structural semantics during the generation process. By encoding primitive semantics into the color domain, the method leverages the strong priors of pre-trained diffusion models to align with well-defined primitives. This enables the inference of multi-view normal maps and semantic maps from a single image, facilitating the reconstruction of a mesh with primitive labels. 
    Furthermore, we introduce geometric optimization techniques and topology-preserving extraction methods to mitigate noise and distortion in the generated primitives. These enhancements result in a complete and seamless B-rep of the CAD model. Experimental results demonstrate that our method effectively recovers high-quality CAD objects from single-view images. Compared to existing 3D generation techniques, the B-rep models produced by CADDreamer are compact in representation, clear in structure, sharp in edges, and watertight in topology.
\end{abstract}    
\section{Introduction}
\label{sec:intro}

Recently, the field of image-based 3D generation has undergone significant advancements~\cite{poole2022dreamfusion, liu2023zero1to3, liu2023one2345, liu2023syncdreamer, shi2023MVDream, hong2023lrm, xu2024instantmesh}. Progress in diffusion models trained on large-scale datasets~\cite{rombach2022stablediffusion, objaverse, hong2023lrm} has greatly streamlined the transformation of 2D images into 3D models, heralding a revolution in 3D content creation.

Despite the increasing amount of work on image-based 3D generation, a critical issue remains. The meshes produced by these methods are typically derived from low-level representations such as neural implicit fields~\cite{mildenhall2020nerf, wang2021neus}, leading to over-tessellated meshes that lack explicit geometric structure and semantics. In contrast, human designers often conceptualize 3D objects at a higher level, perceiving shapes as compositions of basic primitives and feature curves~\cite{lin2020modeling}. As a result, designer-created 3D meshes are compact in representation while preserving clear structures and sharp edges. This significant disparity limits the practical applicability of 3D generative models in scenarios demanding high-quality, structured 3D models, such as gaming, manufacturing, and product design.

To bridge the quality gap between shapes created by human designers and those generated by 3D generative models, this paper focuses on reconstructing Computer-Aided Design (CAD) models from single-view images using diffusion-based generative models. CAD objects offer a higher level of abstraction in shape modeling and ensure structural integrity aligned with human perception. Consequently, generating CAD objects requires diffusion models to understand and interpret high-level structures and geometric primitives. However, single-view images capture only partial information about 3D shapes, necessitating the inference of complete CAD models comprising structured geometric primitives. Additionally, reconstructing the watertight boundary representation (B-rep) of CAD objects poses significant challenges due to inevitable noise and distortion in the generated geometric primitives. Even minor deviations in primitives can disrupt intersection edges, resulting in misaligned primitives or non-watertight B-reps that are unsuitable for manufacturing and product design.

To address these challenges, we propose \emph{CADDreamer}, a method for reconstructing CAD objects from single-view images. It consists of the following two main modules.

\noindent \textbf{Module 1 - Multi-view Generation}: CADDreamer reconstructs a complete 3D mesh and segments it into patches representing six types of geometric primitives: planes, cylinders, cones, spheres, tori, and boundary feature lines. Specifically, we employ a cross-domain multi-view diffusion model~\cite{wonder3d} to jointly predict normal maps and semantic primitive maps. The normal maps are fed into NeuS~\cite{wang2021neus} to reconstruct the complete 3D mesh. Semantic primitive maps are back-projected onto the mesh faces, and a Graph Cut process divides the reconstructed mesh into patches, each corresponding to a geometric primitive.

\noindent \textbf{Module 2 - Geometric and Topological Extraction}: CADDreamer estimates the primitive parameters for each patch and calculates the vertices, edges, and faces of B-rep by determining intersections between primitives. To ensure accuracy, a geometry optimization algorithm corrects noisy primitive parameters, restoring relationships such as parallelism, perpendicularity, collinearity, and intersections. This ensures proper intersection curves are computed. The resulting topological representation, derived from the segmented mesh, guides the intersection of geometric primitives, ultimately producing a watertight B-rep model with accurate CAD vertices, edges, and faces.

The contributions of this paper can be summarized as:
\begin{itemize}[nolistsep,leftmargin=*]
    \item \textbf{CADDreamer Framework}: We propose CADDreamer, a two-module framework for high-quality CAD reconstruction from single-view images. The first module leverages large-scale diffusion models~\cite{rombach2022stablediffusion}, exhibiting superior generative capabilities and a powerful capacity to handle diverse geometric shapes. With cross-view and cross-domain attention mechanisms, the diffusion model ensures high consistency in both geometry and semantics across multiple views. However, the reconstructed mesh from multi-view images may inevitably suffer from distortions and noise. Even with semantic guidance, these inaccuracies can lead to failed computations of geometric primitive intersections, resulting in misaligned primitives or non-watertight B-reps. To address these issues, the second module of CADDreamer is designed to solve the associated geometric and topological challenges.
    \item \textbf{Semantic-enhanced Multi-view 2D Diffusion}: We propose a multi-view 2D diffusion model that perceives high-level semantics by encoding semantic information into the color space. This approach differs from existing methods~\cite{liu2023one2345, liu2023syncdreamer, xu2024instantmesh}, which primarily focus on generating low-level information such as colors and normals. By enforcing strong priors in diffusion models, our method aligns operations with well-defined primitives, inherently enabling the model to understand and interpret high-level structures of CAD objects.
    \item \textbf{Geometry Optimization Algorithm}: We introduce a geometry optimization algorithm to refine the parameters of geometric primitives. This ensures the preservation of topological and geometric relationships between neighboring primitives, such as intersections, parallelism, perpendicularity, and collinearity.
    \item \textbf{Topology-preserving B-rep Construction}: We extract a topological representation from segmented meshes and use it to guide the topology-preserving extraction of CAD vertices, edges, and faces. This process ultimately contributes to generating a watertight and accurate B-rep of CAD objects.
\end{itemize}

\begin{figure*}[htbp] 
    \centering
   \begin{overpic}[width=0.9\linewidth]{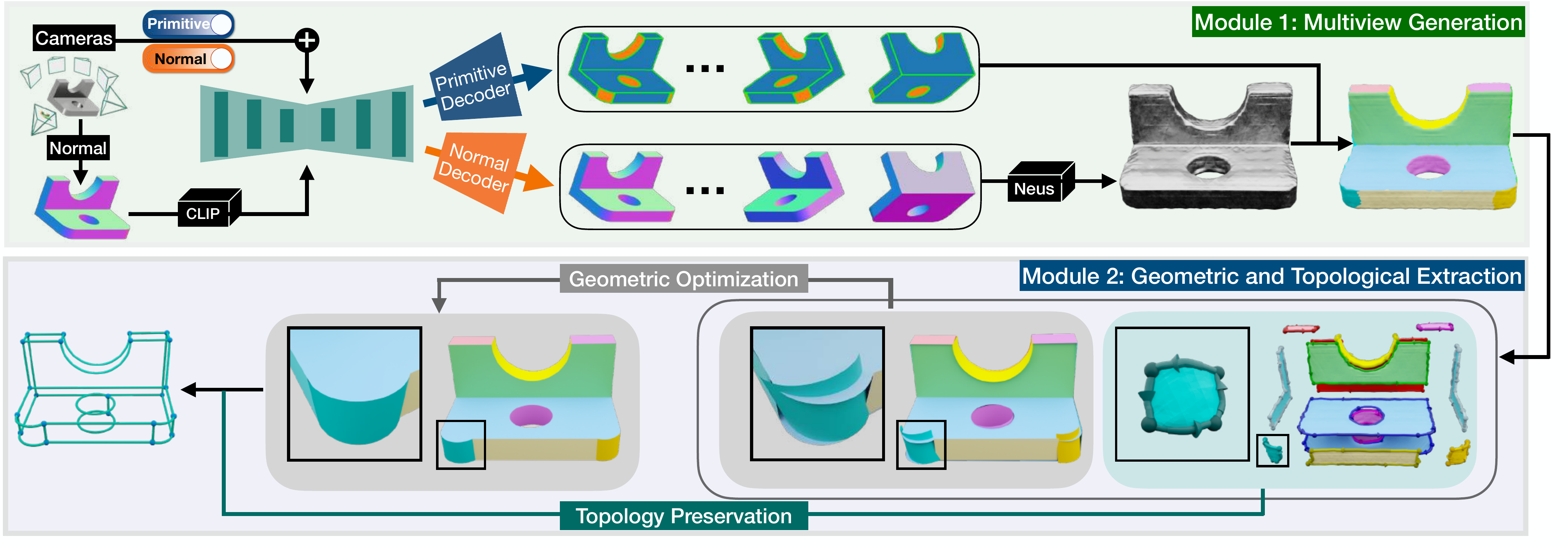}
    \end{overpic}
    \caption{The pipeline of CADDreamer. In the first module, the given single-view RGB image is converted as a normal map. Using the normal map as input, the generation module uses a diffusion process to generate multi-view normal and semantic primitive maps. Inputting the multi-view normal map into Neus~\cite{wang2021neus}, we obtain 3D meshes; back-projecting semantic primitive maps into 3D meshes, we segmented the mesh into several patches with a Graph Cut process. In the second module, geometric optimization corrects the noisy primitive parameters, while the topology-preserving extraction computes their topology-guided intersections and reconstructs a watertight B-rep CAD model.}
    \label{fig:backbone}
\vspace{-2mm}
\end{figure*}

\section{Related work}
\label{sec:rela}

\textbf{Traditional CAD reconstruction methods.} Before the advent of learning-based reconstruction,  a plethora of traditional methods had been developed for B-rep reconstruction, including parameter space-based methods~\cite{limberger2015real,rabbani2007integrated}, 
primitive growing/fitting-based methods~\cite{bauchet2020kinetic,oesau2016planar,li2011globfit,schnabel2007efficient}, and variational surface fitting-based methods~\cite{yan2012variational,zhang2020blending}.  For an overview of these traditional methods, we refer readers to the survey~\cite{kaiser2019survey}.  A major criticism of these traditional methods is that they require users' aid, e.g.,  specifying primitive fitting types manually~\cite{schnabel2007efficient}. Conversely, CADDreamer utilizes a multi-view diffusion module to estimate fitting primitive types and avoid human intervention.

\noindent\textbf{Learning-based CAD Reconstruction Methods}. Recently, with the advancement of deep learning, numerous learning-based CAD generation methods have been proposed. These methods can be divided into three main categories, including  CAD instructions-based methods, retrieval and assembly-based methods, and parameter surface-based methods. Firstly,   CAD instructions-based methods~\cite{wu2021deepcad,li2022free2cad,lambourne2022reconstructing} reduce the  B-rep generation into the natural language generation. By utilizing a large amount of CAD instructions as training data, these methods aim to generate new CAD instructions to achieve CAD generation. However, these methods rely on massive  CAD objects containing CAD instructions, resulting in unacceptable data preparation costs and limiting the potential applications of these methods. Secondly, retrieval and assembly-based methods~\cite{yu2022capri,jayaraman2022solidgen,kania2020ucsg,lambourne2022reconstructing} transform the problem of B-rep reconstruction into a retrieval problem. Using the given images or shapes as keys, these methods retrieve several CAD objects with similar shapes and learn a potential assembly scheme to achieve B-rep reconstruction, i.e., CSG trees~\cite{kania2020ucsg,yu2022capri}. Because maintaining a vast and comprehensive CAD database for retrieving typically incurs massive data preparation costs, these methods are not the first choice for image-to-CAD conversion. Recently, parameter surface-based methods~\cite{li2023surface,liu2023point2cad,xu2024brepgen} have made significant progress in CAD reconstruction. These methods first divide the shape into several parameter surfaces (e.g., primitives), and then estimate the parameters and topologies of these parameter surfaces to implement CAD reconstruction.  These methods have made rapid progress in CAD reverse engineering, but due to the constraints of their pipelines, they cannot take images as inputs and achieve image-to-CAD conversion.

\noindent\textbf{Image2CAD.} Most existing Image2CAD methods rely on a retrieval-and-assembly pipeline based on implicit surfaces rather than directly generating B-reps~\cite{retri1,retri2,retri3,retri4,retri5}. 
Another solution~\cite{chen2024img2cad,alam2024gencad,you2024img2cad} generates B-reps with the conventional sketch-extrude commands.  While these methods can produce B-reps directly, the generated objects are limited to planes and cylinders. In contrast, CADDreamer overcomes these limitations by directly generating B-reps with a diverse set of primitives.

\noindent\textbf{Single-view Reconstruction Methods.}  Large-scale diffusion models~\cite{rombach2022stablediffusion}, pre-trained on billion-scale data, have shown tremendous potential and robust performance in both 2D and 3D generation tasks. Numerous methods based on large-scale diffusion modules have been applied to tackle the problem of single-view 3D reconstruction and have achieved state-of-the-art performance~\cite{xu2024instantmesh,wang2024crm}, such as Wonder3D~\cite{wonder3d}, SyncDreamer~\cite{liu2023syncdreamer}, and LRM~\cite{hong2023lrm}. Despite their success, these approaches typically depend on low-level representations to reconstruct 3D shapes, which lack high-level shape comprehension. This often results in significant noise and distortions in the reconstructed meshes. In contrast, CADDreamer is designed to capture both low-level geometric representations and high-level structured representations of 3D shapes, including seven types of primitives, to mitigate reconstruction noises and contribute to more compact geometric representations.

\section{Method}
\label{sec:method}
As shown in Figure~\ref{fig:backbone}, CADDreamer consists of two main modules: a multi-view generation module and a geometric and topological extraction module. First, the multi-view generation module takes a single-view RGB image to reconstruct a complete 3D triangular mesh and then decomposes this mesh into distinct patches, each corresponding to a primitive. Second, the geometric and topological extraction module refines the parameters of each primitive and performs CAD reconstruction through topology-preserving intersections between neighboring primitives.  
Sec.~\ref{sec:mdiff} details the multi-view generation module, while Sec.~\ref{sec:gom} elaborates on the geometric and topological extraction module.

\begin{figure}[htbp]
    \centering
    \begin{overpic}[width=0.9\linewidth]{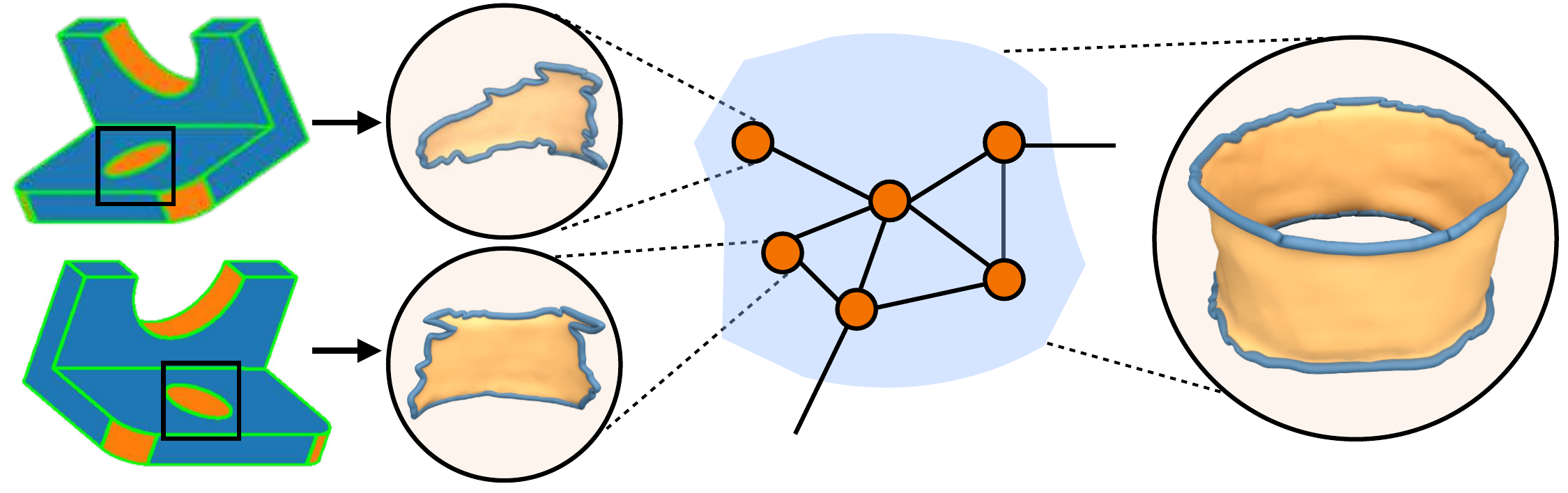}
    \put(12, -2){\textbf{(a)}}    
    \put(65, -2){\textbf{(b)}}    
    \end{overpic}
    \caption{An example of the Graph Cut process to obtain complete 3D mesh patches representing primitives. (a) depicts the mesh patches generated by feature-line cut and back-projection. (b) illustrates the process of Graph Cut algorithm to obtain mesh patches corresponding to primitives. }
    \label{fig:graphcut}
\vspace{-4mm}
\end{figure}
\subsection{Multi-view Generation Module}
\label{sec:mdiff}
The multi-view generation module is designed to reconstruct a complete 3D mesh and then partition the triangles into distinct patches, each representing a CAD primitive. While numerous existing multi-view diffusion modules achieve state-of-the-art performance in 3D mesh reconstruction, these methods devote significant effort to reconstructing multi-view textures and surface materials, which are unnecessary for CADDreamer. Instead, our multi-view generation module focuses on both reconstructing a complete 3D mesh and generating a structured shape representation—specifically, segmenting the mesh into triangle patches that represent CAD primitives. 

Thus, we develop a new multi-view generation module, as illustrated in Figure~\ref{fig:backbone}. The entire process can be summarized in four steps. 
First, we use a popular normal predictor~\cite{bae2024dsine,li2024era3d} to generate a normal map from the single-view RGB image. 
Second, using the generated normal map as input, we fine-tune the popular cross-domain diffusion generative model, Wonder3D~\cite{wonder3d}, to produce $m$ multi-view normal maps and $m$ multi-view semantic primitive maps, where $m=6$ and feature lines have a width of three pixels in this paper. In these maps, the RGB values in the normal maps indicate normal directions, while those in each primitive map represent primitive types, including feature lines that separate neighboring primitives. To enhance generation performance, we fine-tune two separate VAE decoders for normal and primitive map generation, respectively. Third, we input multi-view normal maps into NeuS~\cite{wang2021neus} for 3D mesh reconstruction. We remove the multi-view color inputs and associated texture reconstruction loss from NeuS, as CADDreamer does not require texture reconstruction. Finally, we utilize feature lines to divide each primitive map into several 2D patches. These 2D patches are then back-projected onto the mesh, where a Graph Cut process is applied to merge the back-projected patches into a cohesive set of 3D mesh patches that represent different primitives.

\noindent \textbf{Graph Cut Process.} As shown in Figure~\ref{fig:graphcut}, each back-projected mesh patch is treated as an individual node, and connectivity edges are added between adjacent or overlapping patches. Mesh patches sharing triangle boundaries are considered adjacent, while those with overlaps are overlapped. Using Efficient RANSAC~\cite{schnabel2007efficient}, we extract primitive parameters and calculate their cosine similarity as edge weights. Since the 3D patches derived from the same images correspond to different primitives, merging patches from the same images is considered an error in the graph cut methodology. Given that the number of primitives in CAD objects typically exceeds two, an initial cut threshold is set to divide the mesh into two surface patches. This threshold is then gradually increased to yield more connected components. The minimum cut threshold that results in the fewest errors is selected as the final cut threshold. Thus, through the Graph Cut process, the multi-view generation module effectively separates the constructed mesh into distinct surface patches, each representing a CAD primitive.

Compared to previous models, our multi-view generation module incorporates two significant improvements. First, we introduce an off-the-shelf normal prediction model that converts the original images into normal maps, thereby alleviating the interference caused by image textures and shading variations. As a result, the impact of diverse textures and shading can be minimized, significantly reducing the problem's complexity and contributing to a more robust reconstruction process. Second, we implement a new cross-domain generation strategy for multi-view normal and semantic primitive maps. By generating these maps together, we not only reconstruct a 3D mesh but also segment it into a structured representation of shapes, i.e., the five types of primitives illustrated in  Table~\ref{tab:primitiveparams}.

\begin{table}
    \begin{tabular}{c|l}
        \hline
        Primitives & Parameters \\
            \hline
            \begin{tabular}{c}
            Plane
            \end{tabular} & 
            \begin{tabular}{l}
            $\vec{x}, p$  : normal and position\\
            \end{tabular} \\
        \hline
            \begin{tabular}{c}
            Cylinder
            \end{tabular} & 
            \begin{tabular}{l}
            $\vec{x}, p$  : axis  and position, \\
            $r$ : radius
            \end{tabular} \\
        \hline
            \begin{tabular}{c}
            Cone
            \end{tabular} & 
            \begin{tabular}{l}
            $\vec{x}, p$  : axis and center position, \\
            $\alpha$ : semi-angle,  $h$ : height
            \end{tabular} \\
        \hline
            \begin{tabular}{c}
            Torus
            \end{tabular} & 
            \begin{tabular}{l}
            $\vec{x}, p$  : axis and center position \\
            $r_l$ : major radius, $r_s$ : minor radius
            \end{tabular} \\
        \hline
            \begin{tabular}{c}
            Sphere
            \end{tabular} & 
            \begin{tabular}{l}
            $p$ : center position, $r$ :  radius
            \end{tabular} \\
        \midrule[1pt]
        \hline
        Relationships & Determination \\  
        \hline
            \begin{tabular}{c}
            Topology \\
            Relationship
            \end{tabular} & 
            \begin{tabular}{l}
            Intersected : ( Intersection is not  $\emptyset$)
            \end{tabular} \\
        \hline
        \begin{tabular}{c}
            Geometric\\
            Relationships
            \end{tabular} & 
            \begin{tabular}{l}
                Perpendicular: ($\vec{x_1} \cdot \vec{x_2} = 0$)\\
                Parallel: ($\vec{x_1} \cdot \vec{x_2} = 1$)\\
                Collinear : ($\exists t: p_1 = p_2 + t\vec{x_2}$)
            \end{tabular} \\
        \hline
    
    \end{tabular} \\
    \caption{Definitions of parameters and relationships for primitives. }
    \label{tab:primitiveparams}
\vspace{-4mm}
\end{table}

\subsection{Geometric and Topological Extraction Module}
\label{sec:gom}
The multi-view generation module in Sec.\ref{sec:mdiff} converts the given image into a 3D mesh and decomposes it into multiple mesh patches, each representing a distinct primitive. The geometric optimization module then aims to extract primitive surfaces and generate a CAD model by finding intersections between these primitives. This process consists of four main steps: 
\begin{itemize}
    \item First, we input the vertices of each mesh patch into the RANSAC~\cite{schnabel2007efficient} algorithm to extract the parameters for each primitive, as defined in Table~\ref{tab:primitiveparams}. 
    \item Next, we introduce a novel geometric optimization algorithm to recover the topological relationships (e.g., intersections) and geometric relationships (e.g., parallelism, perpendicularity, and collinearity) between primitives, with detailed definitions provided in Table~\ref{tab:primitiveparams}.
    \item Third, we extract a topological representation from the segmented mesh patches, encompassing vertices, edges, and faces, as referred to Figure~A1 in the supplement.
    \item Finally, using the extracted topological representation, we perform topology-preserving intersection operations between each primitive surface and its neighboring primitives to create CAD models' vertices, edges, and faces. 
\end{itemize}

The details of each step of the process are outlined below.

\begin{figure}[htbp]
    \centering
    \includegraphics[width=1.0\linewidth]{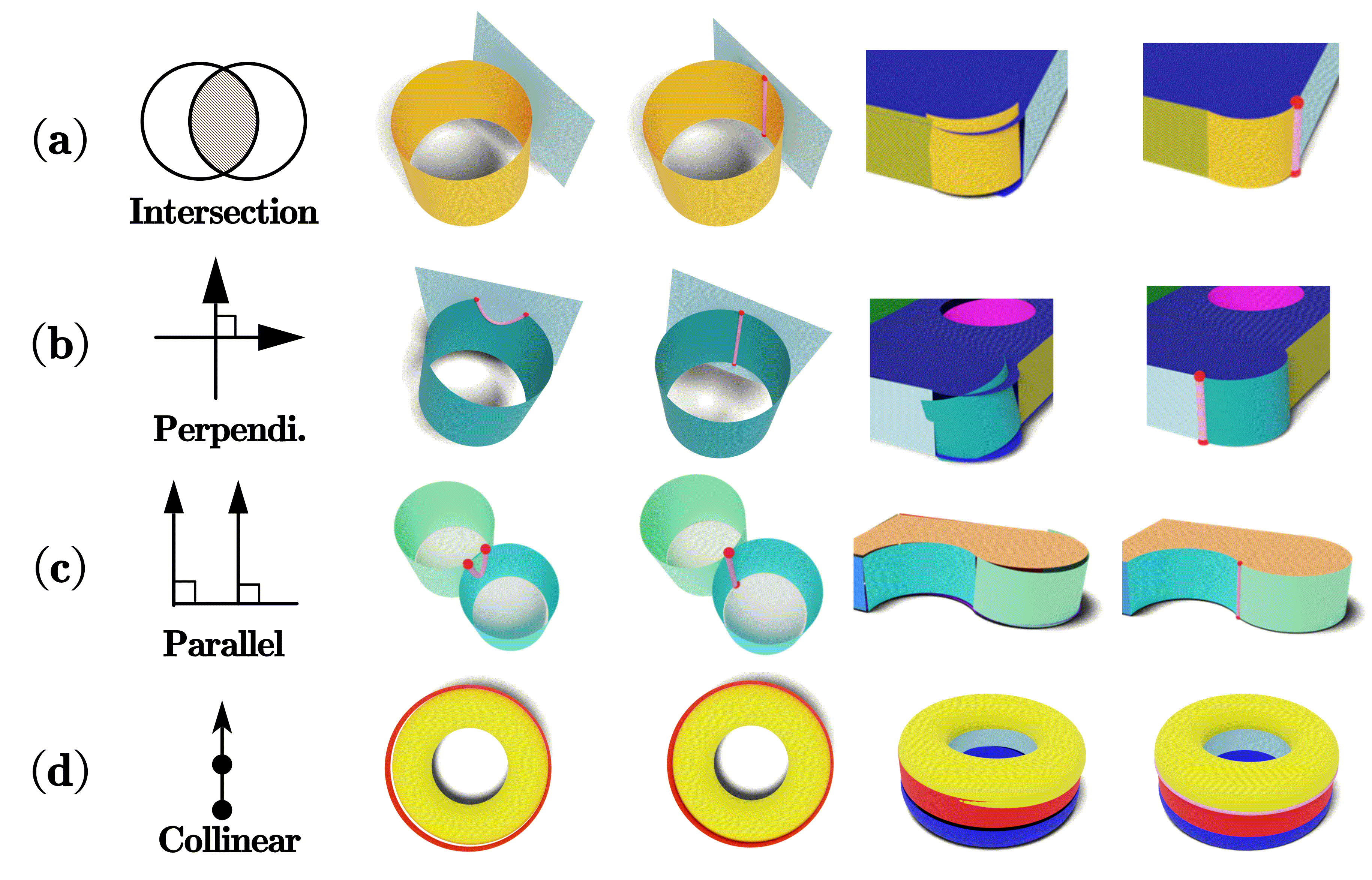}
    \caption{ Four key primitive relationships: (a-b) cylinder-plane intersection and perpendicularity; (c) parallel cylinders; (d) collinear cylinder and torus. Incorrect relationships (columns 2 \& 4) yield flawed intersections, while correct ones (columns 3 \& 5) produce accurate results. }
    \label{fig:different_relationship}
\vspace{-4mm}
\end{figure}

\noindent \textbf{Primitive Extraction.} In the first step, we build on previous studies~\cite{schnabel2007efficient} by using the RANSAC algorithm to estimate the parameters of primitives. Unlike direct least squares fitting, which can be significantly affected by outlier data, RANSAC helps to mitigate the impact of such outliers. However, due to the inevitable distortion in 3D mesh reconstruction from single-view images, the primitive parameters obtained from RANSAC may still be inaccurate. As a result, some essential topological and geometric relationships between the primitives might not be preserved. As illustrated in Figure~\ref{fig:different_relationship}(a), the expected intersecting primitive surfaces may fail to intersect, leading to incorrect topological relationships and reconstruction failures for vertices and edges. Furthermore, since the parameters of primitives are based on points, lines, and directions, it is crucial to preserve key geometric relationships between these elements, such as collinearity among points and lines, as well as parallelism and perpendicularity among directions. As shown in Figure~\ref{fig:different_relationship}(b), (c), and (d), neglecting these vital geometric relationships can result in completely erroneous intersection curves, which may cause gaps in the reconstructed CAD model. Therefore, in the next step, we aim to establish these important relationships between primitives before implementing the intersection reconstruction strategy.

\begin{figure*}[htbp]
    \centering
    \begin{overpic}[width=\linewidth]{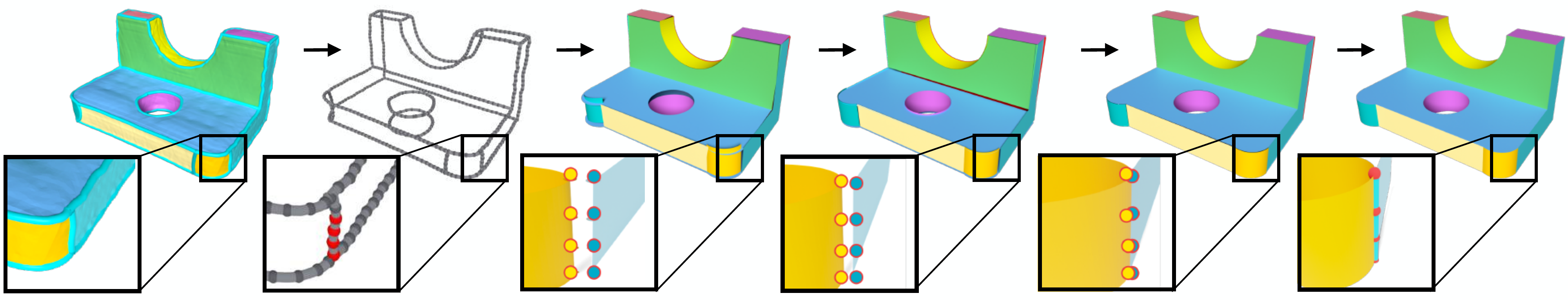}
    \put(3, -1){\textbf{(a)}}
    \put(20, -1){\textbf{(b)}}
    \put(36, -1){\textbf{(c)}}
    \put(53, -1){\textbf{(d)}}
    \put(70, -1){\textbf{(e)}}
    \put(86, -1){\textbf{(f)}}
    \end{overpic}
    \vspace{-10pt}
    \caption{An example of primitive stitching process with four stitching vertices. (a) shows the patch boundaries on reconstructed mesh, (b) represents the sampled stitching vertices (k=4), (c) illustrates the initial primitives and the projected points of stitching vertices, (d) depicts the stitching result after the first step of optimization, (e) shows the result at the 100th step, and (f) represents the final stitching result.   }
    \label{fig:stitching}
\vspace{-4mm}
\end{figure*}
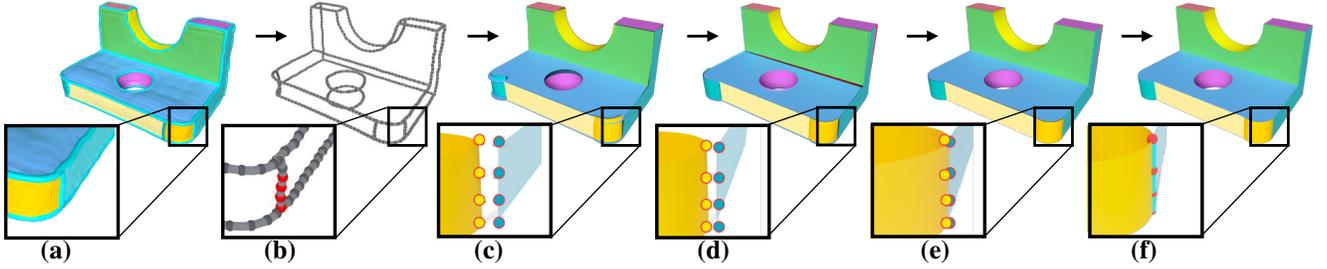

\noindent \textbf{Geometric Optimization.} In the second step, we introduce a novel geometric optimization algorithm designed to refine primitive parameters and restore four types of primitive relationships. Specifically, our algorithm includes a primitive stitching process aimed at recovering intersection relationships while incorporating geometric relationships as constraints to ensure that the final primitive parameters conform to these geometric relationships. The primitive stitching process can be divided into three key sub-steps.

First, we derive primitive intersection relationships from adjacent patches on segmented meshes: primitives intersect if and only if neighboring patches share boundaries. Second, for each intersection relationship, we sample $k$ boundary vertices closest to intersecting primitives as \emph{stitching vertices} (Figure~\ref{fig:stitching}(b)). 
We uniformly sample candidate points along mesh segmentation boundaries (line-strings) and select $k$ points with minimal projection distances to the intersecting primitives as stitching vertices.
Third, we project each stitching vertex onto the two primitive surfaces and minimize the distance between the corresponding projection points by optimizing the parameters of the primitive surfaces, as shown in Figure~\ref{fig:stitching}(c-f).

The optimization function for a stitching vertex $v_i$ between the two corresponding primitive surfaces $A$ and $B$ is defined as follows:
\begin{equation}
    f_{stch}(v_i) = || \pi(v_i, P_A) - 
        \pi(v_i, P_B) ||,
\end{equation}
where $P_A$ and $P_B$ are the optimizable shape parameters of the primitive surfaces $A$ and $B$, and $\pi(v_i, P_A)$ and $\pi(v_i, P_B)$ represent the projection points of $v_i$ on $A$ and $B$, respectively. By minimizing the sum of optimization functions for all stitching vertices $\sum_{v_i \in \mathcal{V}}f_{stch}(v_i)$, the primitive stitching algorithm reduces the distances between primitive surfaces that should intersect, helping to prevent intersection failures and preserve correct topological relationships between neighboring primitives. 

We also incorporate certain geometric relationships as constraints within the primitive stitching process to ensure that the resulting parameters satisfy these relationships. 
\begin{itemize}
    \item A \emph{parallel} relationship means that two primitives share the same axis direction, allowing us to optimize a single axis direction while maintaining consistency between the axes of both primitives.
    \item A \emph{collinear} relationship implies that the position of primitive $A$ (denoted as $p_A$) can be expressed in terms of the parameters of primitive $B$: $p_A = p_B + \vec{x}_B * t$, where $\vec{x}_B$ is the axis direction of primitive $B$, as shown in Table~\ref{tab:primitiveparams}. Therefore, to enforce collinearity constraints, we optimize the parameter $t$ instead of $p_A$.
    \item For a \emph{perpendicular} relationship, we utilize the dot product between the two axes as an additional optimization function. Given the set of all stitching points $\mathcal{V}$ and all perpendicular relationships $\mathcal{P}$, the  optimization function can be expressed as:
    $\sum_{v_i \in \mathcal{V}} f_{stch}(v_i) + \sum_{(C, D) \in \mathcal{P}} \vec{x}_C \cdot \vec{x}_D$, where $C$ and $D$ are perpendicular primitives, and $\vec{x}_C$ and $\vec{x}_D$ are their axis directions, respectively.
\end{itemize}
In our implementation, the three geometric relationships are detected if the equations from Table~\ref{tab:primitiveparams} hold within a tolerance of 0.05. By employing the L-BFGS algorithm~\cite{fei2014parallel} to optimize the objective function, we effectively minimize the distances between primitive surfaces that should intersect and reconstruct the geometric relationships between primitives. Figure~\ref{fig:stitching} illustrates an example of four stitching points. As shown in Figure~\ref{fig:stitching} (d), by enforcing constraints to restore parallel and collinear relationships, the first step of the optimization process re-establishes the parallel relationship between the axes of the cylinders and planes. Subsequently, as the optimization progresses, we gradually restore the perpendicular and intersection relationships between the primitives, as shown in Figure~\ref{fig:stitching} (e-f).

\noindent \textbf{Extraction of Topological Representation.} In the third step, we extract a watertight topological representation from the mesh to guide the intersection of the primitives. Specifically, each 3D mesh patch corresponds to a topological face, while the edge curves adjacent to two patches are considered topological edges. Additionally, mesh vertices connected to more than two patches are regarded as topological vertices. The half-edge structure of the meshes provides directionality for the patch boundaries, and we inherit these directions as the orientations of the topological edges. This process enables us to extract a topological representation from the segmented mesh, including topological vertices, edges, and faces (see Figure~A1 in the supplementary material).  Since the reconstructed meshes are watertight, this topological representation is also watertight.

\begin{figure*}[htbp]
    \centering
    \includegraphics[width=1.0\textwidth, trim=0.4cm 0.4cm 0.4cm 0.4cm, clip]{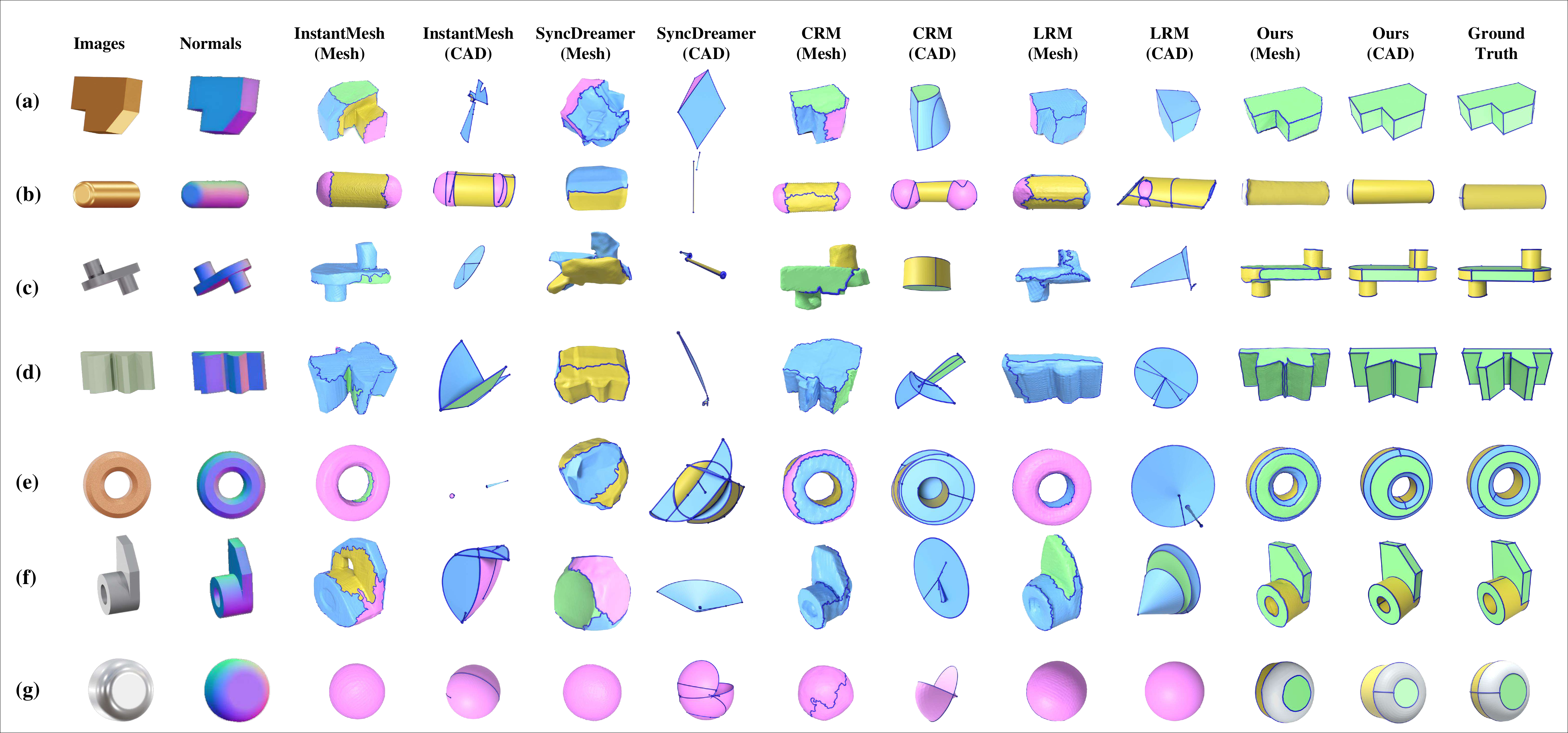}
    \caption{The segmentation results and reconstructed B-rep models. Electric blue (\tikzcircle[black, fill=customRed]{4pt}) represents segmentation boundaries and edges of B-rep objects. Other colors represents different primitives:  plane (\tikzcircle[black, fill=green]{4pt}), cylinder (\tikzcircle[black, fill=customBlue]{4pt}), cone (\tikzcircle[black, fill=customYellow]{4pt}), sphere (\tikzcircle[black, fill=customMagenta]{4pt}), and torus (\tikzcircle[black, fill=customCyan]{4pt}).}
    \label{fig:exp_1}
\vspace{-2mm}
\end{figure*}
\noindent \textbf{Topology-preserving CAD Reconstruction.} In the final step, we utilize this watertight topology as guidance to construct the CAD model using the following intersection strategy, which is divided into two main sub-steps. 

First, we reconstruct the CAD curves for each face. 
Referring to Figure~A1(a) in the supplement, each topological edge's neighborhood comprises two patches corresponding to two primitives. We then compute the intersection of these two primitives to obtain CAD curves. If multiple intersection curves exist between two primitives, we select the one with the closest projection distance to the corresponding topological edge. By applying this process to all topological edges, we generate the CAD curves for each one.

Second, we reconstruct the CAD vertices, edges, and faces using the following intersection strategy. Each topological vertex connects two topological edges, or CAD curves. We compute the intersection of these two corresponding CAD curves as the intersection point of the three related primitives, given that each CAD curve is associated with two neighboring primitives and one primitive is shared between the two CAD curves. If multiple intersected vertices are present, we select the vertex closest to the topological vertex as the reconstructed vertex. By applying this process to all topological vertices, we obtain an equal number of CAD vertices. Using these CAD vertices to trim the CAD curves allows us to create CAD edges. Furthermore, by utilizing these CAD edges to trim the primitives, we obtain CAD faces. By merging these CAD vertices, edges, and faces, we create a complete and watertight boundary representation (B-rep) for the CAD model.

\section{Experiments}

\textbf{Dataset.} We begin by training and evaluating CADDreamer on synthetic images, followed by testing it on real-world images. For the synthetic dataset, we curate 30,000 seamless CAD models from the ABC~\cite{Koch2019ABC} and DeepCAD~\cite{wu2021deepcad} datasets. These CAD objects are enhanced with textures and backgrounds, then rendered into multi-view images, resulting in a training set of 29,000 samples and a testing set of 1,000 samples. To maintain consistency, we adopt the camera configurations and image resolution settings used in Wonder3D~\cite{wonder3d}. Furthermore, we capture RGB images of real CAD objects using handheld devices to assess our model's generalization performance on real-world objects.

\noindent\textbf{Baselines.} Currently, popular approaches for reconstructing B-reps fall into two categories: conventional sketch-extrude methods~\cite{chen2024img2cad,alam2024gencad,you2024img2cad} and more recent intersection-based methods~\cite{liu2023point2cad}. 
Because our dataset includes CAD models with a diverse range of primitives—such as spheres, tori, and cones—that cannot be reconstructed using conventional sketch-extrude methods~\cite{chen2024img2cad,alam2024gencad,you2024img2cad}, we focus on comparing intersection-based models, which can handle these varied primitive types. 
Our evaluation framework integrates state-of-the-art single-view reconstruction methods (e.g., SyncDreamer~\cite{liu2023syncdreamer}, LRM~\cite{hong2023lrm}, CRM~\cite{wang2024crm}, and InstantMesh~\cite{xu2024instantmesh}) with B-rep reconstruction techniques like Point2CAD~\cite{liu2023point2cad}. For benchmarking, we apply these single-view generation methods to create meshes, and then use Point2CAD to segment the meshes into primitive shapes, estimate primitive parameters, and convert them into B-reps.  Additional comparative assessments of the traditional sketch-extrude approach within point cloud inputs are documented in the supplementary material.

\begin{table}[t]
    \centering
    \resizebox{\linewidth}{!}{
    \begin{tabular}{lcccc}
    \toprule
    \textbf{Methods} &  \textbf{CD ($\downarrow$) } 
                        & \textbf{NC ($\uparrow$)}
                     & \textbf{SEG(V) ($\uparrow$)}
                     & \textbf{SEG(P) ($\uparrow$)}   \\ \hline
    CRM~\cite{wang2024crm} & 3.97 & 64.4 & 40.2 & 49.3  \\
    LRM~\cite{hong2023lrm} & 4.26 & 63.6 & 38.4 & 46.8    \\
    InstantMesh~\cite{xu2024instantmesh} & 4.61 & 58.3 & 35.1 & 41.7   \\ 
    SyncDreamer~\cite{liu2023syncdreamer} & 5.49 & 48.9 & 29.8  & 33.2  \\ 
    CADDreamer & \textbf{1.27} & \textbf{92.6} & \textbf{95.7} & \textbf{97.9} \\ 
    \bottomrule
    \end{tabular}
    }
    \caption{
            Statistical results of reconstructed meshes and extracted primitives, including Chamfer distance (CD $\times 100$),  normal consistency (NC, \%), segmentation accuracy based on vertices (SEG(V), \%),  and segmentation accuracy based on primitive counts (SEG(P),\%). Best values are highlighted in bold. 
    } 
    \label{tab:meshexp}
\vspace{-5mm}
\end{table}
\noindent\textbf{Metrics.}  We evaluate generation quality across three aspects: meshes, primitives, and B-reps. First, an ideal reconstructed mesh should minimize reconstruction distortion and align closely with the original B-reps (i.e., primitives). To assess geometric alignment between the reconstructed mesh and the ground truth, we utilize two widely-used metrics: Chamfer Distance (CD) and Normal Consistency (NC)~\cite{gkioxari2019mesh}. Second, incorrect segmentation results—particularly the omission of primitives—can cause significant fitting errors and reconstruction failures. To address this, we evaluate both the percentage of correctly classified mesh vertices (SEG(V)) and the accuracy of predicted primitive counts (SEG(P)). Additionally, fitting errors in primitives can disrupt correct intersection relationships, resulting in hanging faces or non-closed B-reps. To quantify these issues, we calculate the proportion of hanging faces (HF) in the resulting B-reps and measured the deviation of the reconstructed B-reps from the ground truth using Chamfer Distance (CD).

\noindent\textbf{Settings.}  We use Wonder3D~\cite{wonder3d} as our pre-trained model to ensure more stable and faster convergence. All experiments are conducted on a single machine equipped with eight NVIDIA A100 GPUs (80GB each) and an AMD EPYC 7313 CPU. The details of the comparative analyses are presented below, while ablation studies and limitation analyses are provided in the supplementary materials.

\begin{table}[t]
    \centering
    \resizebox{0.7\linewidth}{!}{
    \begin{tabular}{lcc}
    \toprule
    \textbf{Methods}  &  \textbf{HF ($\downarrow$)   }
                     &  \textbf{CD   ($\downarrow$)}\\ \hline
    CRM~\cite{wang2024crm} & 35.2 & 9.74   \\
    LRM~\cite{hong2023lrm} & 39.6  & 11.6    \\
    InstantMesh~\cite{xu2024instantmesh} & 43.6 & 13.1    \\ 
    SyncDreamer~\cite{liu2023syncdreamer} & 58.5  & 15.4   \\ 
    CADDreamer & \textbf{2.4}  & \textbf{1.36}   \\ 
    \bottomrule
    \end{tabular}
    }
    \caption{
            Statistical results of reconstructed B-reps, including the percentage of B-reps with hanging faces (HF, \%),  and Chamfer distance (CD $\times 100$) between the reconstructed B-reps and ground truth. Best values are highlighted in bold.   
    }
    \label{tab:cadexp}
\vspace{-5mm}
\end{table}

\subsection{Comparisons on Synthetic Images}
\label{sec:exp_1} 
 Figure~\ref{fig:exp_1} visually showcases the reconstruction results from synthetic images, while Table~\ref{tab:meshexp} and~\ref{tab:cadexp} provide quantitative comparisons. These results demonstrate that CADDreamer offers significant advantages in mesh generation quality, primitive extraction, and the topological fidelity of the reconstructed B-reps.

CADDreamer's key strength lies in its ability to simultaneously capture low-level geometric features (normal maps) and high-level semantic shape understanding (primitive maps), resulting in meshes with minimal distortion.  As shown in Figure~\ref{fig:exp_1}, our method generates clean, compact meshes with sharp edges, minimizing fitting errors and reconstruction failures. In contrast, competing methods, which lack high-level semantic understanding, generate distorted reconstructions that significantly deviate from the primitives. 
Quantitative results in Table~\ref{tab:meshexp} further validate CADDreamer's performance, showing that its meshes exhibit the least geometric deviation from the ground truth, with the lowest CD and highest NC values.

In addition, CADDreamer employs a back-projection and Graph Cut process for primitive extraction, rather than relying on traditional mesh or point cloud segmentation methods. This approach enables more precise primitive extraction. Single-view RGB image reconstructions often produce meshes with errors such as uneven surfaces and over-smoothed boundaries. When these flawed meshes are processed through segmentation-based methods for primitive extraction, the reconstruction errors are amplified. 
For example, two twisted planes with a smooth transition might be misidentified as a single cone, as shown in Figure~\ref{fig:exp_1}(a) LRM. Using such incorrectly identified primitives for mesh reconstruction inevitably results in excessive fitting errors and reconstruction failures. 
In contrast, our method demonstrates superior primitive extraction that aligns closely with ground truth CAD models. Our segmentation approach achieves more precise primitive labeling than other methods, with $97.9\%$ of meshes correctly matching the primitive count of the ground truth, as shown in Table~\ref{tab:meshexp}.

Thirdly, CADDreamer refines critical relationships between adjacent primitives, such as intersection, parallelism, 
perpendicularity, and collinearity. 
It also employs a topology-preserving intersection strategy to reconstruct B-reps with accurate topology, reducing two common errors: hanging faces that arise from incorrect primitive intersections, and massive non-closed B-reps caused by non-intersection, as shown in Figure~\ref{fig:exp_1}(a-f). As shown in Table~\ref{tab:cadexp}, CADDreamer minimizes the percentages of hanging faces (HF) while significantly reducing face reconstruction failures caused by non-intersection, resulting in the lowest CD among compared approaches.
\begin{figure}[tbp]
    \centering
    \includegraphics[width=1.0\linewidth, trim=0.4cm 0.4cm 0.4cm 0.4cm, clip]{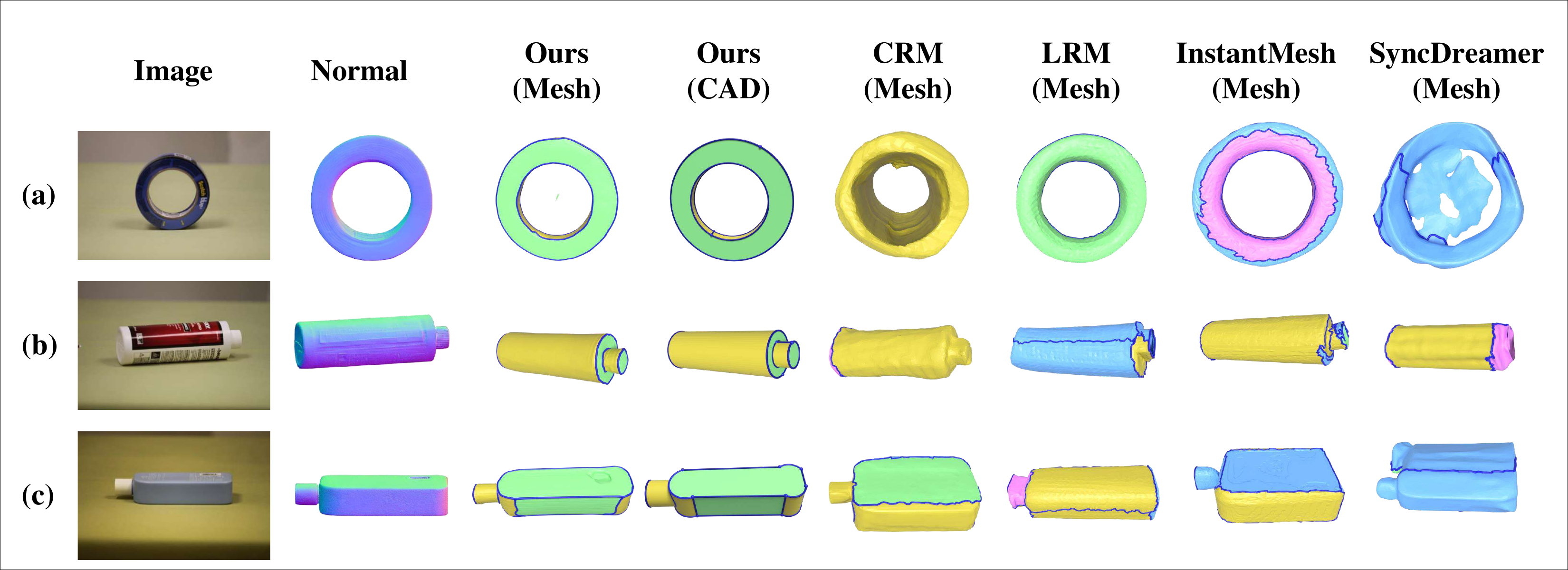}
    \caption{Segmentation results and reconstructed B-reps from real images (color definitions are the same as in Figure~\ref{fig:exp_1}). }
    \label{fig:exp_2}
\vspace{-5mm}
\end{figure}
\subsection{Comparisons on Real Images}
\label{sec:exp_2}
We capture RGB images of real-world CAD objects using a Canon EOS R5 camera with a Canon EF 70 lens. Era3D~\cite{li2024era3d} is employed for background removal and as a normal predictor to generate accurate normal maps, which are subsequently used as inputs to evaluate our method's generalization capability on real-world CAD objects. As shown in Figure~\ref{fig:exp_2}, 
despite the fact that real images possess more complex lighting and shadow effects, greater perspective distortion, and introduce larger errors in normal estimation, our method still successfully reconstructs high-quality CAD models, which demonstrates the generalization potential of our approach in handling real images.

\section{Conclusion and Future Work}
CADDreamer introduces a novel cross-domain, multi-view generation strategy, along with a geometric optimization and topological extraction process, enabling the accurate generation of B-reps with diverse primitives. However, two key limitations remain. First, the method's performance is constrained by image quantity and resolution, sometimes failing to detect extremely fine geometric features, which leads to fitting inaccuracies and incomplete reconstructions. Second, like existing single-view reconstruction approaches, the system struggles with challenging viewpoints and complex occlusions, especially when the captured angles do not reveal all the primitives. Our future work will explore multi-view generation techniques that incorporate additional viewpoints and enhanced resolution to address these limitations.


{
   \small
   \bibliographystyle{ieeenat_fullname}
   \bibliography{main}
}
\clearpage

\setcounter{page}{1}
\setcounter{table}{0}
\setcounter{figure}{0}
\renewcommand{\thetable}{A\arabic{table}}
\renewcommand{\thefigure}{A\arabic{figure}}

\maketitlesupplementary
\appendix
\begin{figure}[htbp]
    \centering
    \begin{overpic}[width=\linewidth]{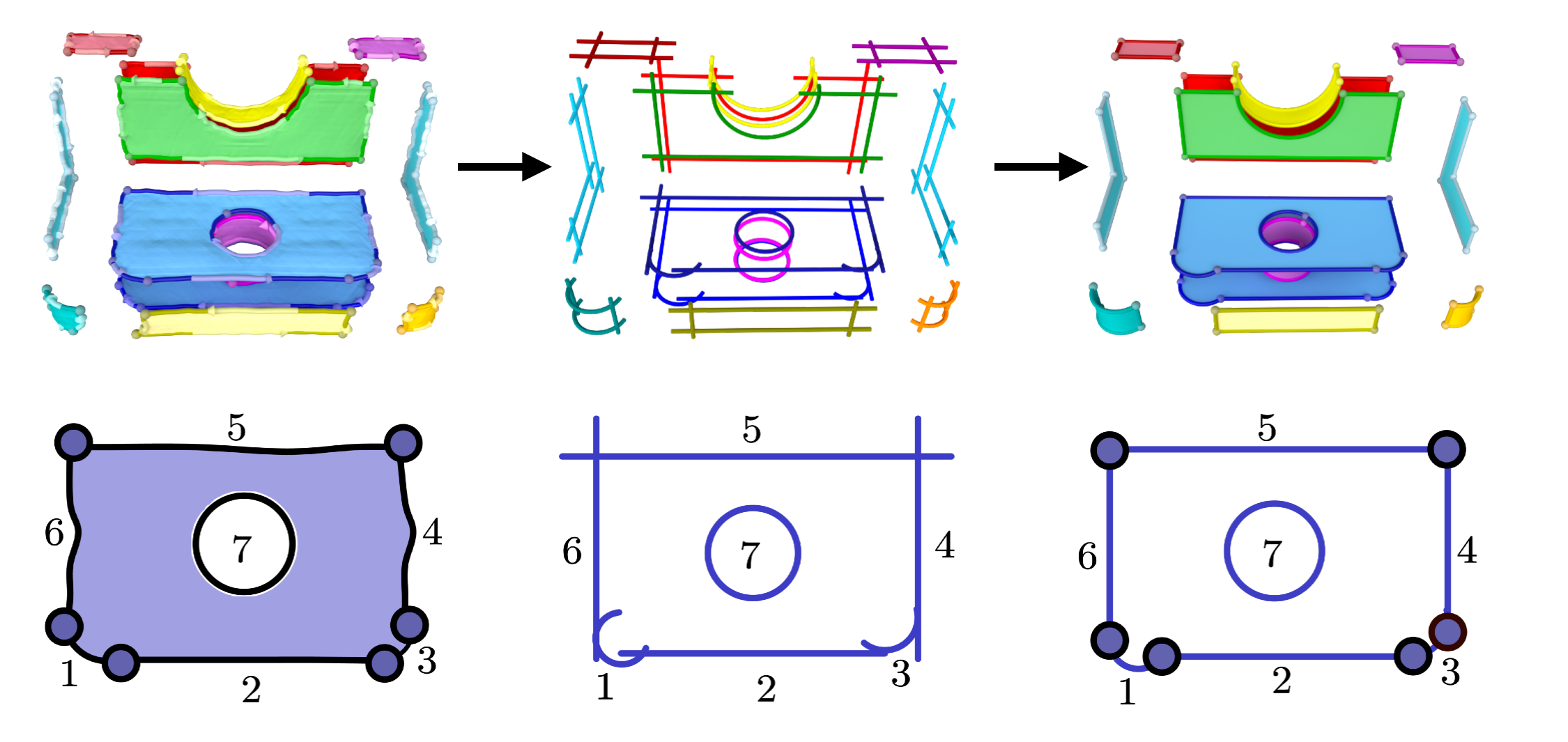}
    \put(13, -2){\textbf{(a)}}    
    \put(45, -2){\textbf{(b)}}    
    \put(80, -2){\textbf{(c)}}    
    \end{overpic}
    \vspace{-10pt}
    \caption{An illustration of the topology-preserving intersection strategy to achieve a watertight B-rep model with CAD vertices, edges, and faces. (a) depicts the topology derived from the mesh patches. (b) shows the CAD curves resulting from primitive intersections. (c) represents the reconstructed B-rep model through topology-preserving trimming.  }
    \label{fig:intersection}
\vspace{-4mm}
\end{figure}
Our project page and code can be found \href{https://github.com/lidan233/CADDreamer}{here}.
In this supplementary document, we provide more details about the evaluation metrics used in this paper, the ablation study, our limitations, and failure cases, as well as more experimental results.

\section{Details of Evaluation Metrics}
In this section, we provide more details about each evaluation metric used in this paper.
\begin{itemize}
    \item \textbf{Chamfer Distance (CD).} We uniformly sample 10,000 points from both the generated and ground truth CAD models. The Chamfer Distance is the average minimum distance between these point pairs.
    \item \textbf{Normal Consistency (NC).} We sample 10,000 points uniformly from both the generated and ground truth CAD models. Normal consistency is then computed as the average absolute cosine similarity between corresponding point pairs, identified via nearest neighbor search.
    \item \textbf{Vertex-wise Segmentation Accuracy (SEG(V)).} For each case, 10,000 points are uniformly sampled from both the ground truth and the segmented mesh/CAD model, along with their primitive labels. SEG(V) is the average percentage of points whose nearest neighbor in the ground truth has a matching label. 
    \item \textbf{Primitive-wise Segmentation Accuracy (SEG(P)).} For each case, we directly compare the number of primitives in the segmentation result. If the number of segmentation labels is correct, we consider it a correct prediction.  SEG(P) is the percentage of cases with correct predictions.
    \item \textbf{Ratio of Hanging Faces (HF).}  For each case, if a face has an edge with no neighboring face, we consider it a hanging face. We calculate the ratio between the number of hanging faces and the total number of all faces. HF is the average of these ratios across all cases.
\end{itemize}
\begin{table}[ht]
    \centering
    \resizebox{\linewidth}{!}{
    \begin{tabular}{lcccc}
    \toprule
    \textbf{Methods} &  \textbf{CD ($\downarrow$) } 
                        & \textbf{NC ($\uparrow$)}
                     & \textbf{SEG(V) ($\uparrow$)}
                     & \textbf{SEG(P) ($\uparrow$)}   \\ \hline
     Model 1  & 2.56 & 80.3 & 58.8 & 67.8  \\
     Model 2  & 1.28 & 92.4 & 63.8 & 69.4  \\
     Model 3  & 1.27 & 92.5 & 95.5 & 97.6  \\
     CADDreamer & \textbf{1.27} & \textbf{92.6} & \textbf{95.7} & \textbf{97.9} \\ 
    \bottomrule
    \end{tabular}
    }
    \caption{
            Statistical results of reconstructed meshes and extracted primitives across ablation models. Evaluation metrics include: Chamfer distance (CD ×100), normal consistency (NC, \%), vertex-based segmentation accuracy (SEG(V), \%), and primitive count-based segmentation accuracy (SEG(P), \%). Best values are highlighted in bold.
    } 
    \label{tab:ablamesh}
\vspace{-5mm}
\end{table}
\begin{table}[ht]
    \centering
    \resizebox{0.5\linewidth}{!}{
    \begin{tabular}{lcc}
    \toprule
    \textbf{Methods}  &  \textbf{HF ($\downarrow$)   }
                     &  \textbf{CD   ($\downarrow$)}\\ \hline
    Model 1    &  32.4  &  8.84 \\
    Model 2    &  29.8  &  8.13 \\
    Model 3    &  2.40  &   1.37 \\
    Model 4    & 23.5 &  6.47 \\
    Model 5    & 11.8  &  3.29 \\
    CADDreamer & \textbf{2.4}  & \textbf{1.36}   \\ 
    \bottomrule
    \end{tabular}
    }
    \caption{
            Statistical results of reconstructed B-reps across ablation models, 
            including the percentage of B-reps with hanging faces (HF, \%),  
            and Chamfer distance (CD $\times 100$) between the reconstructed B-reps and ground truth. 
            Best values are highlighted in bold.   
    }
    \label{tab:ablacad}
\vspace{-5mm}
\end{table}

\section{Ablation Study}
Our method benefits primarily from two distinct technical contributions: a new multi-view generation module that yields higher-quality meshes and improved segmentation results, and a geometric optimization strategy that reconstructs essential topological and geometric relationships to ensure CAD reconstruction through primitive intersection. We conduct ablation studies on these two technical contributions to investigate the performance gains achieved independently by each module.

\begin{figure}[tbp]
    \centering
    \includegraphics[width=1.0\linewidth, clip]{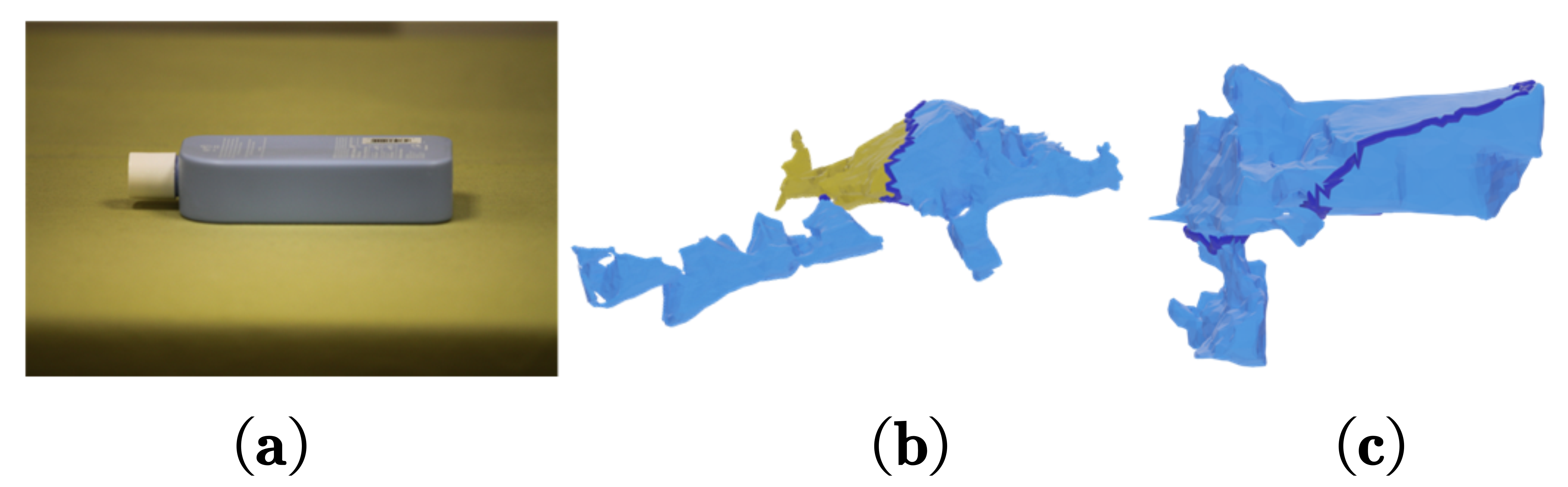}
    \caption{Using RGB images as input for fine-tuning will lead to poor generalization performance on real images. (a)  real image input, (b-c) are the segmentation results of Model 2-3, which are fine-tuned with RGB inputs. }
    \label{fig:abla_real_img}
\vspace{-2mm}
\end{figure}
\begin{figure}[tbp]
    \centering
    \includegraphics[width=1.0\linewidth, clip]{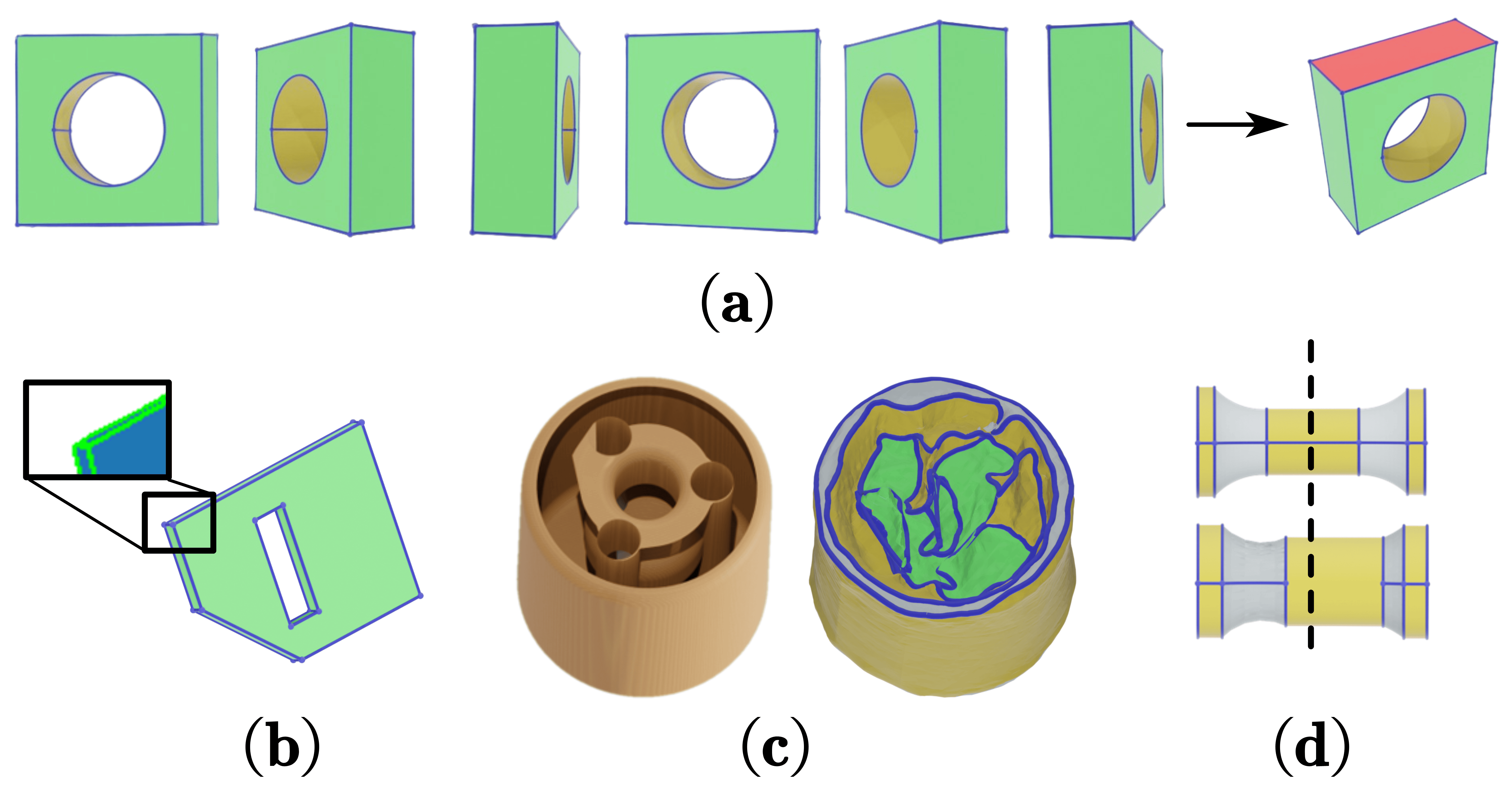}
    \caption{The limitations of our methods. (a)  The multi-view images of six horizontal angles cannot cover the top face (red) of the CAD object. (b) An extremely thin and elongated plane (less than 1 pixel wide). (c) A failure case of complex geometry (original image and segmentation result). (d) Our generated B-reps cannot preserve symmetry of the CAD structure. }
    \label{fig:limitation}
\vspace{-2mm}
\end{figure}
\textbf{Generation module.} We first evaluate the contributions of our generation module through three comparative models. Model 1 employs the non-finetuned Wonder3D (RGB inputs and RGB\&normal outputs) combined with Point2CAD to establish a baseline for comparison with other fine-tuned solutions. Model 2 utilizes the fine-tuned Wonder3D (RGB inputs and RGB\&normal outputs) with Point2CAD to assess the effectiveness of our new cross-domain generation strategy (normal\&semantic map). Model 3 implements CADDreamer's pipeline with RGB inputs to evaluate the impact of our normal inputs. As demonstrated in Table~\ref{tab:ablamesh}, all models fine-tuned on primitive-based shapes outperform the original Wonder3D (Model 1), which lacks prior knowledge of primitives. Models that incorporate our cross-domain generation strategy for mesh segmentation (Models 3-5) achieve superior segmentation results compared to the post-processing segmentation approach applied to noisy meshes (Models 1-2), leading to more accurate B-reps reconstruction outcomes. Although using RGB images as inputs (Model 3) does not diminish CADDreamer's performance on synthetic data, it impairs generalization to real-world image inputs, as illustrated in Figure~\ref{fig:abla_real_img}. This limitation stems from the training data's restricted texture variety and fixed lighting conditions, which fail to capture the complexity and variability of textures and illumination encountered in real-world scenarios.

\textbf{Geometry optimization module.} Next, we investigate the contribution of our geometric optimization strategy and the importance of enforcing parallelism, perpendicularity, and collinearity constraints. We construct two variant models: (1) \emph{Model 4}, which removes all geometric optimization steps from CADDreamer, and (2) \emph{Model 5}, which maintains primitive stitching but removes the three geometric constraints. Since Models 4 and 5 share the same mesh reconstruction and segmentation results with our method, we focus our analysis on comparing the reconstructed B-reps with ground truth, as shown in Table~\ref{tab:ablacad}. Given that our method provides mesh and segmentation results closer to ground truth, the hanging faces and CD metrics of Models 4 and 5 demonstrate a significant improvement over Point2CAD (Model 1-2). Without geometric optimization, reconstruction fitting errors introduce gaps between primitives, compromising geometric relationships and causing missing or incorrect intersections (see Figure~4 in the main manuscript).   Model 5, incorporating primitive stitching, effectively reduces intersection failures, thus achieving lower hanging face ratios and CD values compared to Model 4. Additionally, our model with three geometric constraints reduces reconstruction failures caused by erroneous geometric relationships, resulting in fewer hanging faces and smaller Chamfer distances than Model 5.

\section{Limitations and Failure Cases}
CADDreamer consists of two main modules: the generation module and the geometry optimization module. We discuss the limitations of these two modules separately.

\textbf{Generation module.} Constrained by limited multi-view angles and image resolutions, the generation module fails to maintain robustness under extreme viewing angles and struggles with CAD shapes containing fine geometric structures.  First, CADDreamer generates mesh segmentation schemes using multi-view images, requiring the selected viewpoints to include all CAD faces. Since our method uses Wonder3D as a pretrained model with the same multi-view setup (six 256×256 images), our method requires that the CAD faces must be included in these six viewpoints.  As illustrated in Figure~\ref{fig:limitation}(a), when the six viewpoints used for generation are all horizontal, the top face of the CAD object does not appear in any of the six views. Consequently, the top face is omitted in the final generation, leading to inaccurate topology and reconstruction failure.  Second, some B-reps contain extremely thin and elongated CAD faces that, when projected into multi-view images, manifest as sub-pixel features with widths less than one pixel. As demonstrated in Figure~\ref{fig:limitation}(b), the elongated top face is segmented into isolated pixels by feature lines, which is not a continuous patch and cannot generate meaningful segmentation results. The combination of these limitations means that CADDreamer might fail to reconstruct and segment complex CAD shapes, especially shapes with fine structures and excessive occlusion, as shown in Figure~\ref{fig:limitation}(c). A potential solution to these limitations is to increase the number of views and improve image resolution: adding a top view ensures coverage of the top face, while using higher image resolution  makes fine structures more visible, enabling better face coverage. However, this approach demands prohibitive computational resources and introduces greater convergence challenges.

\textbf{Geometry optimization module} only reconstructs basic geometric relationships between primitives, such as parallelism, perpendicularity, and collinearity. It does not consider more complex relationships between multiple primitives, such as symmetry. As shown in Figure~\ref{fig:limitation}(d), while the ground truth (upper) exhibits perfect symmetry between primitives, such property is difficult to be preserved in our results (lower) due to inevitable reconstruction and fitting errors. Furthermore, since detecting and optimizing such complex relationships remains an open problem, implementing their reconstruction within the geometric optimization module presents significant challenges.

\section{More Comparison Results}
We add additional comparative experiments to analyze the advantages of CADDreamer. First, we add the comparison with DeepCAD~\cite{wu2021deepcad} to analyze the advantages of CADDreamer compared to sketch-extrude methods. In addition, we notice some closed-source (Tripo~\cite{tochilkin2024triposr}) and open-source (Trellis~\cite{xiang2024structured}, MeshAnything~\cite{chen2024meshanything}) models released concurrently with CADDreamer. We analyze the performance of these methods on some typical examples and explain why these methods are not suitable for CAD generation.
\begin{figure}[tbp]
    \centering
    \includegraphics[width=1.0\linewidth, trim=2 4 2 2, clip]{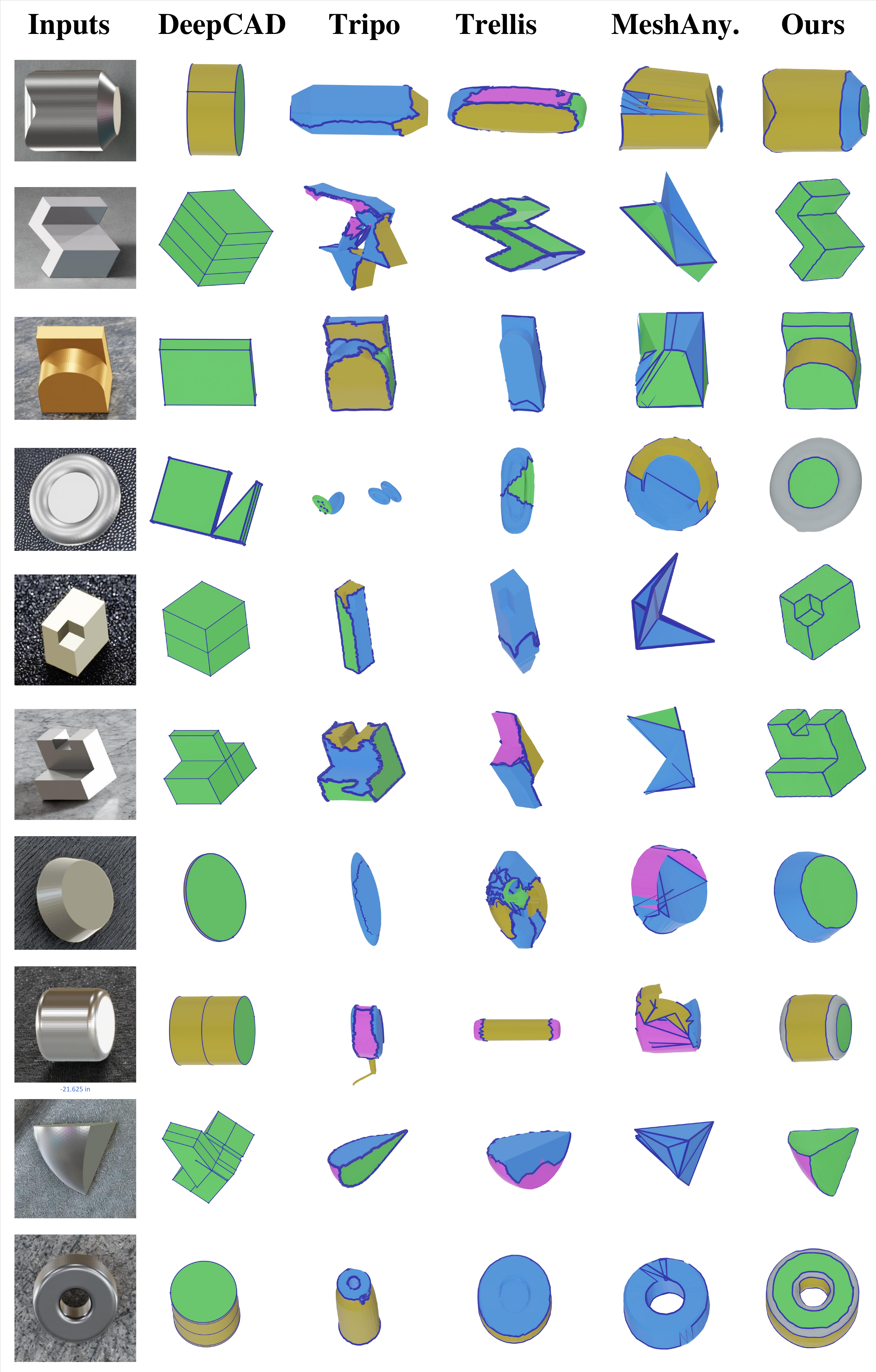}
    \caption{Comparison with other methods. From left to right, (a) Input image, (b) DeepCAD results~\cite{wu2021deepcad}, (c) Tripo results~\cite{tochilkin2024triposr}, (d) Trellis results~\cite{xiang2024structured}, (e) MeshAnything results~\cite{chen2024meshanything}), and (f) our segmentation results. Color definitions match those used in the main manuscript figures.}
    \label{fig:other_comparisions}
\end{figure}

Since DeepCAD cannot handle the image inputs, we use our generated meshes as inputs and use DeepCAD to generate B-reps. As shown in Figure~\ref{fig:other_comparisions}, since sketch-extrude approaches can only support planes and cylinders, complex primitives like cones are simplified, and the reconstructed B-reps have higher errors (CD=4.21, NC=67.8) than CADDreamer (CD=1.36, NC=91.9) on our testing dataset.  

We also evaluate Tripo, Trellis, and MeshAnything on 10 diverse CAD models for mesh reconstruction, followed by Point2CAD for CAD generation to obtain CAD models. Tripo and Trellis tend to add unnecessary details and components due to their focus on generating diverse and complex shapes, which results in a large number of segmented primitives and faces, as shown in Figure~\ref{fig:other_comparisions}. MeshAnything, based on a finite-length autoregressive generation progress, overlooks strong constraints on mesh connectivity. Consequently, it tends to generate simple and topology-breaking meshes, resulting in numerous isolated primitives and inferior outcomes than CADDreamer.

\section{More Reconstruction Results}
\begin{figure*}[htbp]
    \centering
    \includegraphics[width=1.0\textwidth, clip]{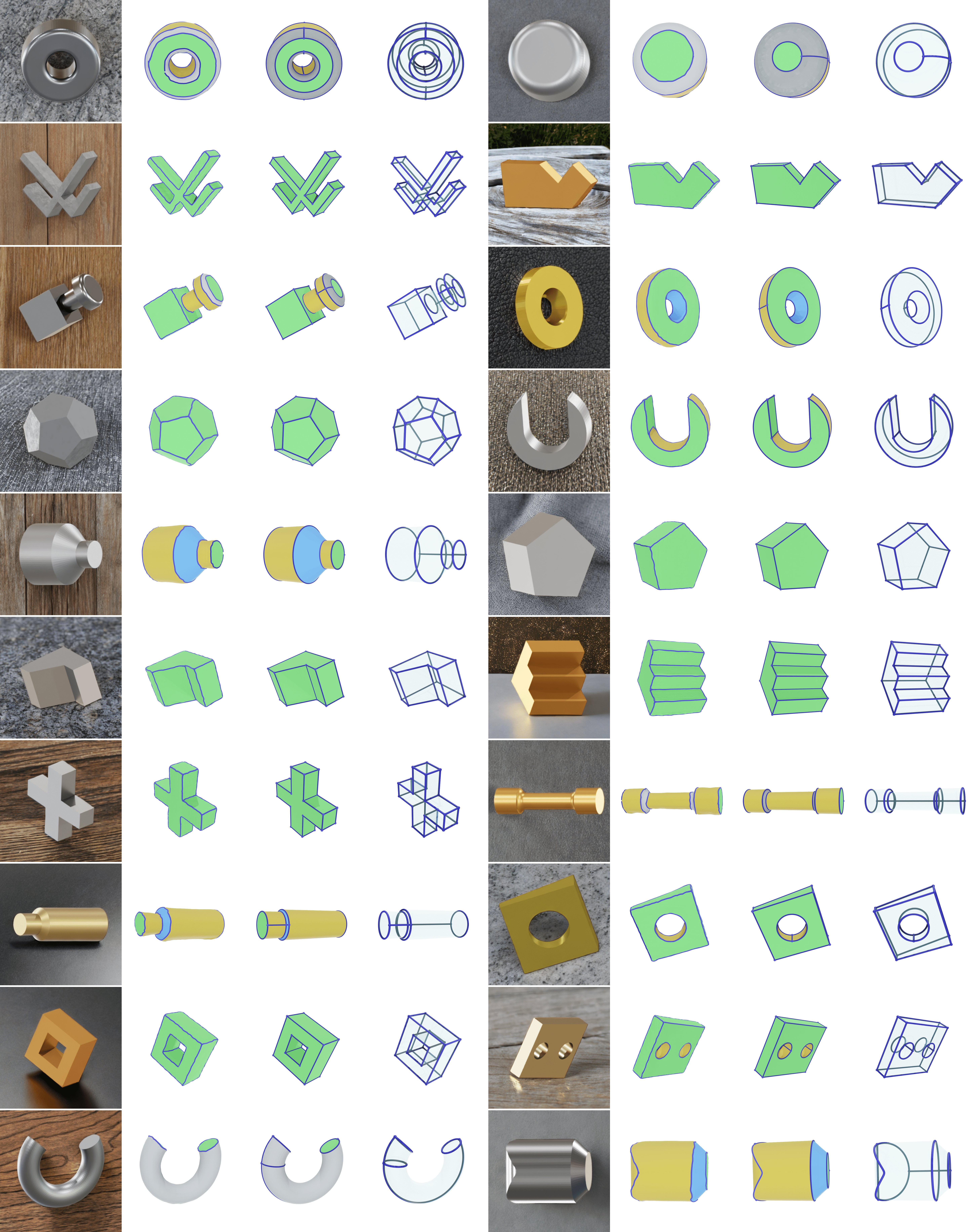}
    \caption{Reconstruction results from the given images are shown from left to right: input image, reconstructed mesh, B-reps, and their CAD vertices and edges. Different colors indicate different primitives: plane (\tikzcircle[black, fill=customGreen]{4pt}), cylinder (\tikzcircle[black, fill=customBlue]{4pt}), cone (\tikzcircle[black, fill=customYellow]{4pt}), sphere (\tikzcircle[black, fill=customMagenta]{4pt}), and torus (\tikzcircle[black, fill=customCyan]{4pt}).}
    \label{fig:exp22}
    \vspace{-2mm}
\end{figure*}

\begin{figure*}[htbp]
    \centering
    \includegraphics[width=1.0\textwidth, clip]{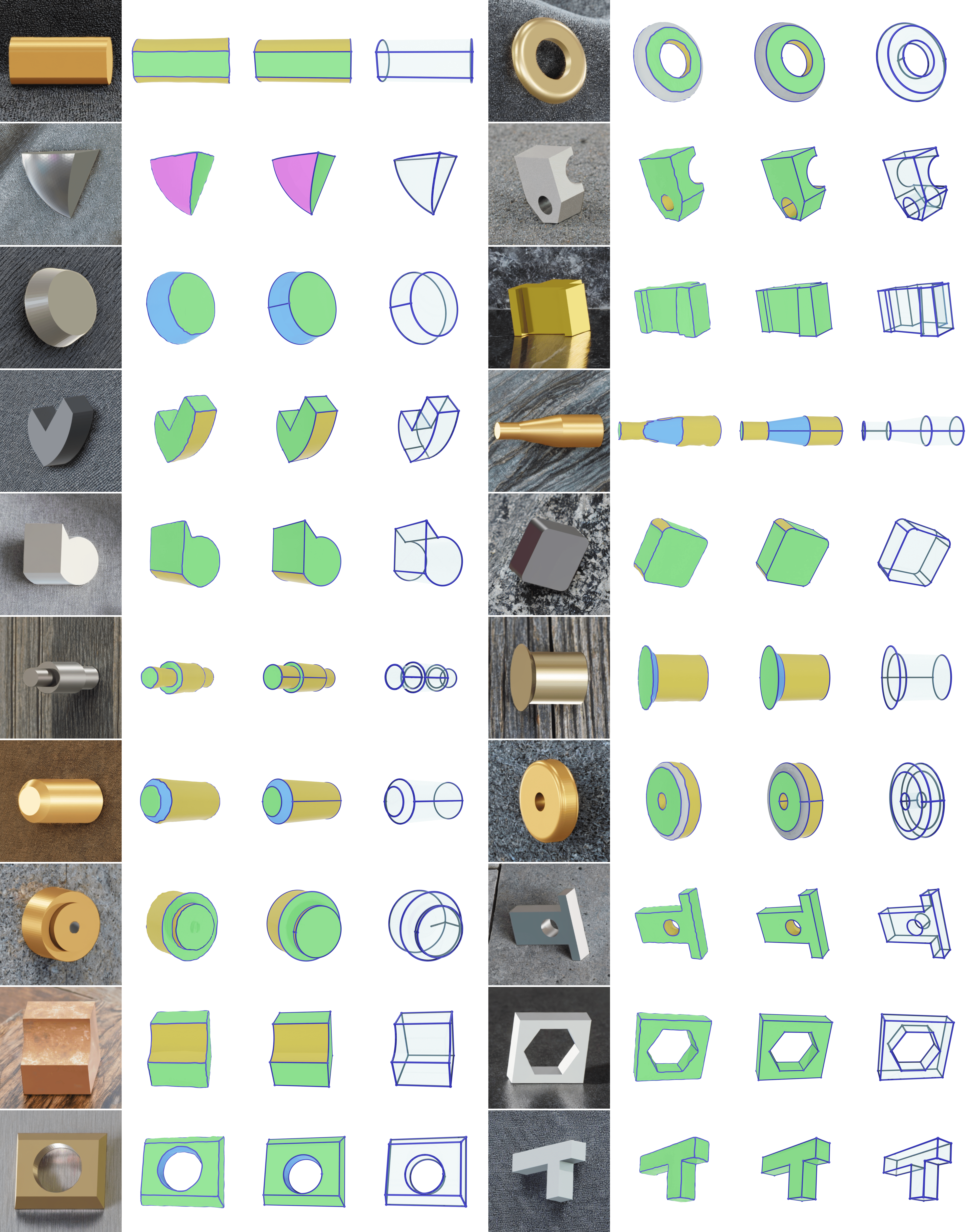}
    \caption{Reconstruction results from the given images are shown from left to right: input image, reconstructed mesh, B-reps, and their CAD vertices and edges. Different colors indicate different primitives: plane (\tikzcircle[black, fill=customGreen]{4pt}), cylinder (\tikzcircle[black, fill=customBlue]{4pt}), cone (\tikzcircle[black, fill=customYellow]{4pt}), sphere (\tikzcircle[black, fill=customMagenta]{4pt}), and torus (\tikzcircle[black, fill=customCyan]{4pt}).}
    \label{fig:exp32}
    \vspace{-2mm}
\end{figure*}

\begin{figure*}[htbp]
    \centering
    \includegraphics[width=1.0\textwidth, clip]{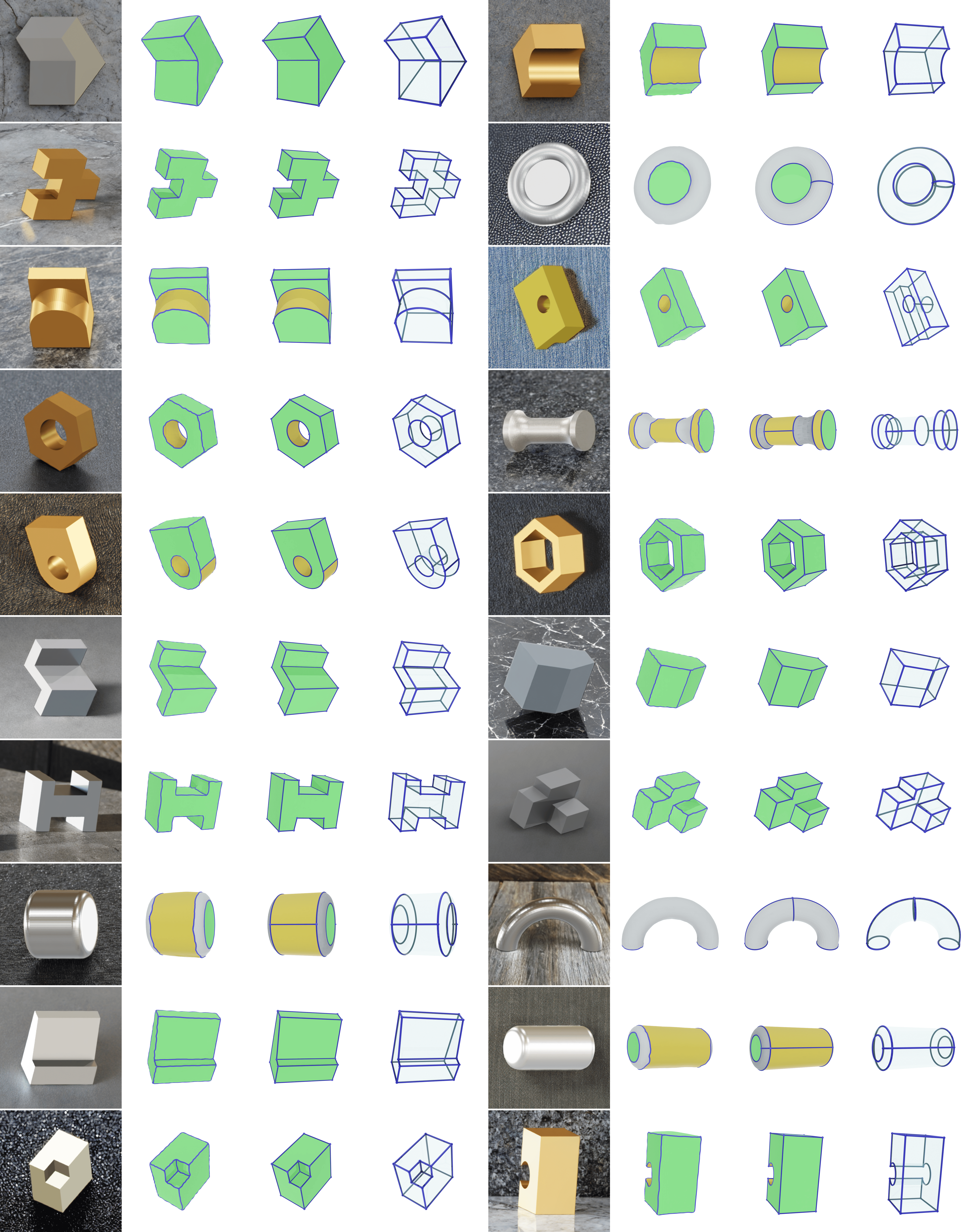}
    \caption{Reconstruction results from the given images are shown from left to right: input image, reconstructed mesh, B-reps, and their CAD vertices and edges. Different colors indicate different primitives: plane (\tikzcircle[black, fill=customGreen]{4pt}), cylinder (\tikzcircle[black, fill=customBlue]{4pt}), cone (\tikzcircle[black, fill=customYellow]{4pt}), sphere (\tikzcircle[black, fill=customMagenta]{4pt}), and torus (\tikzcircle[black, fill=customCyan]{4pt}).}
    \label{fig:exp43}
    \vspace{-2mm}
\end{figure*}

Figure~\ref{fig:exp22},~\ref{fig:exp32}, and~\ref{fig:exp43} showcase more reconstruction results of our method, including input images, reconstructed meshes, reconstructed B-reps, as well as their CAD vertices and edges. 

\end{document}